\documentclass[10pt,twocolumn,letterpaper]{article}

\usepackage{authblk}
\makeatletter
\renewcommand\AB@affilsepx{ \quad \protect\Affilfont}
\makeatother
\usepackage[pagenumbers]{cvpr} % To force page numbers, e.g. for an arXiv version

\definecolor{cvprblue}{rgb}{0.21,0.49,0.74}
\usepackage[pagebackref,breaklinks,colorlinks,allcolors=cvprblue]{hyperref}

\def\paperID{} % *** Enter the Paper ID here
\def\confName{CVPR}
\def\confYear{2026}

\title{Zero-shot Reconstruction of In-Scene Object Manipulation from Video}

\author[1]{Dixuan Lin$^*$}
\author[2]{Tianyou Wang$^*$}
\author[1]{Zhuoyang Pan}
\author[1]{Yufu Wang$^\dag$}
\author[1]{Lingjie Liu}
\author[1]{Kostas Daniilidis}
\affil[1]{University of Pennsylvania}
\affil[2]{University of Oxford}

\begin{document}

\twocolumn[{%
\renewcommand\twocolumn[1][]{#1}%
\maketitle
\begin{center}
    \vspace{-2em}
    \captionsetup{type=figure}
    \includegraphics[width=1\linewidth]{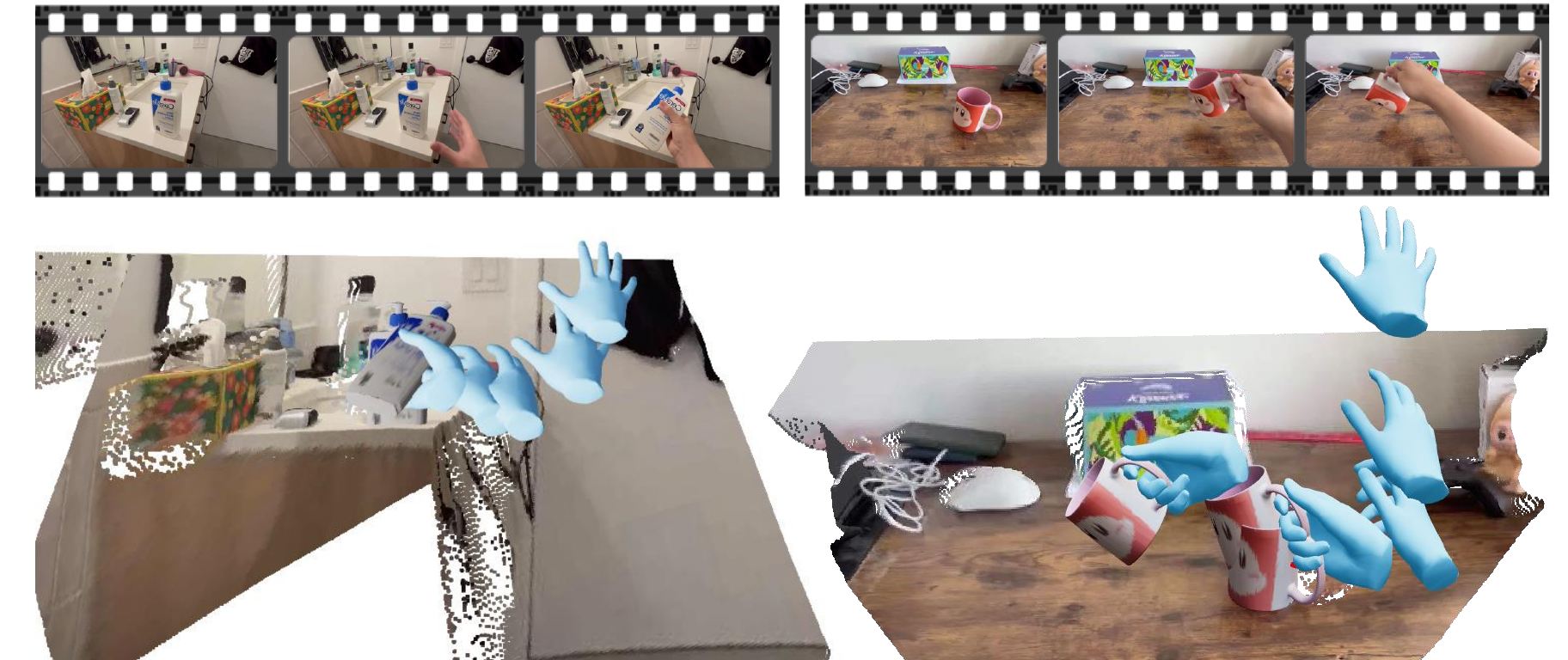}
    \vspace{-1.5em}
    \captionof{figure}{
    We present a zero shot system that reconstruct in-scene object manipulation motion from daily videos.
    }
    \label{fig:teaser}
\end{center}
}]
\maketitle
\begin{abstract}
We build the first system to address the problem of reconstructing in-scene object manipulation from a monocular RGB video. It is challenging due to ill-posed scene reconstruction, ambiguous hand–object depth, and the need for physically plausible interactions. Existing methods operate in hand-centric coordinates and ignore the scene, hindering metric accuracy and practical use. In our method, we first use data-driven foundation models to initialize the core components, including the object mesh and poses, the scene point cloud, and the hand poses. We then apply a two-stage optimization that recovers a complete hand–object motion from grasping to interaction, which remains consistent with the scene information observed in the input video. Project Page: \href{https://reisom2025.github.io/Reisom.github.io/}{reisom2025.github.io/Reisom.github.io/}.
\end{abstract}    
\section{Introduction}
\label{sec:intro}
Reconstructing scene-aligned object manipulation from in-the-wild monocular videos is both practical and impactful. In policy learning for robots with dexterous hands, a policy must learn not only hand–object interactions, but also how the object is used and how it interacts within the surrounding scene. Likewise, AR/VR applications require consistent alignment among the hand, object, and scene to enable reliable and intuitive control. 

% However, the problem remains inadequately solved. Although numerous studies have proposed methods for reconstructing hand–object grasping from a single image and hand–object interaction from a video, these approaches are largely limited to basic actions such as grasping and twisting. They cannot handle task-oriented manipulations—for example, “placing an apple into a basket” or “lifting a kettle and pouring water into a cup.” Motivated by this gap, we aim to develop a system that reconstructs manipulation of unseen objects in unseen scenes.

However, this problem remains largely unexplored, primarily due to three key challenges: (i) Scene reconstruction from a monocular video is inherently ill-posed. Both classical multi-view geometry pipelines (e.g., COLMAP with MVS\cite{schonberger2016sfm,schonberger2016mvs}) and optimization-based neural scene representations (e.g., NeRF~\cite{mildenhall2021nerf}, 3D Gaussian splatting\cite{kerbl20233dgs}) fundamentally require multi-view coverage. (ii) Reconstructing physically plausible grasps and interactions with objects in random shapes remains difficult: methods that perform well on dataset templates often fail to achieve contact-consistent, physically valid interactions in the wild. (iii) Estimating accurate 3D motion of the hand and object within the scene is challenging. Recent hand pose estimation methods~\cite{hamer,haptic,yu2023acr,li2022intaghand} achieve high pose accuracy and strong 2D reprojection consistency, but still struggle to predict accurate 3D depth.

Given the above challenges, existing object manipulation reconstruction methods\cite{diffhoi,hold,magichoi} focus only on scene-agnostic interactions, without reconstructing the surrounding scene. Consequently, tasks such as "Placing an apple into the basket on table" or "Lifting a kettle and pouring water into a cup" remain largely unaddressed.

Motivated by this gap, we develop a system that reconstructs object manipulations in scene coordinates. We decompose the problem into sub-tasks and leverage strong specialist models that effectively address each sub-problem to obtain reliable intermediate results. We then obtain a unified final result through an optimization procedure. We address these three main challenges by designing the following components of our system.
\\
\textbf{Scene reconstruction.} Recent feed-forward approaches \cite{wang2025vggt} show that data-driven priors alone can produce strong reconstructions, preserving accurate geometry without hand-crafted constraints and generalizing to previously unseen objects and scenes. We leverage priors from scene reconstruction methods to generate a scene point cloud from a single monocular view.
\\
\textbf{Hand-object interaction.} We first obtain realistic object meshes utilizing the power of learning-based mesh recovery methods~\cite{xu2024instantmesh,liu2024meshformer,xiang2025trellis} which learned strong priors from large 3D asset corpora. Subsequently, we employ a 2D-to-3D contact point matching algorithm to identify corresponding contact points on hand and object, and optimize the motion to obtain a physically plausible hand–object interaction. 
\\
\textbf{Motion detection and recovery.} The hand’s high degrees of freedom and deformability make metric depth estimation via 2D–3D correspondence fundamentally unreliable. In contrast, shifting the focus to the object greatly simplifies the problem. Using object reconstruction methods, we obtain a textured 3D shape that we model as a rigid body; this enables robust 2D feature matching and recovery of its 6-DoF pose trajectory. Once the object’s pose is localized in the scene, we use it as an anchor to align the hand trajectory, yielding a consistent, scene-aligned motion.

% Building on this insight, we propose a zero-shot in-scene object manipulation reconstruction system that integrates the capabilities of foundation models to reconstruct hand–object manipulation motions consistenting with the scene’s 3D geometry, by inputting only a single-view RGB video. With the aid of the aforementioned foundation models, we obtain a scene point cloud, estimated camera parameters, object shape, a coarse object trajectory, and hand poses. Given these inputs, we focus on addressing following issues:\\
% \textbf{Non-holding stages processing.} Prior hand–object reconstruction works typically consider only the holding phase. In real in-scene tasks, however, we must separately account for the phases where the hand moves in the scene without the object and where the hand manipulates the object.
% \\
% \textbf{Hand object motion alignment.} For holding stage, to ensure a consistent hand–object motion, we require trajectories that are highly coherent in the scene coordinate frame. We address this by optimizing for a globally aligned hand–object motion.
% \\
% \textbf{Continuity from discrete observations. }Heavy occlusions during hand–object interaction cause hand-pose estimators to fail on certain frames; moreover, frames where the hand leaves the camera’s field of view provide no observations. We employ a human motion prior to transform these sparse, discrete hand observations into a temporally continuous pose sequence.

The main contributions of this work are as follows: We introduce the first system that reconstructs scene-aligned hand–object manipulation from monocular RGB video. We demonstrate robust performance on standard benchmarks as well as in-the-wild videos.

\section{Related Works}
\label{sec:related_works}

\subsection{Hand Pose Estimation}
Recovering accurate 3D hand pose underpins understanding of human actions and interactions, and has enabled downstream applications such as robotic manipulation\cite{okami,robotseerobotdo,zeromimic,vlmimic,HOP}. Given the ease of acquiring RGB data and the need for scalable deployment in real-world applications, there has been growing work on estimating 3D hand pose from an RGB image\cite{frankmocap,hamer,jiang2023probabilistic,li2024hhmr,lin2021mesh,park2022handoccnet,prakash20243d,wilor,yu2023acr,zhang2019end,zhou2024simple} or monocular video\cite{Dyn-hamr,haptic,omnihands,harp,hmp,h2onet}. Image-based methods focus on per-frame reconstruction, enforcing consistency with observations, and improving robustness under occlusion. The representative work is HaMeR\cite{hamer}, which scales the training data and adopts a large ViT backbone to achieve strong performance on in-the-wild images. However, these approaches predict hand poses in a canonical space, and depth along the viewing direction is unconstrained, introducing ambiguity and scale mismatch; this limits accuracy and deployment for downstream usability. Our method adopts HaMeR for grasping-frame hand initialization and performs a subsequent optimization to enforce metric-correct depth along each sequence. Video-based methods leverage temporal continuity to enforce coherent and stable 3D hand pose trajectories across frames. Recent work Dyn-HaMR\cite{Dyn-hamr} presents an optimization-based pipeline, which integrates camera motion estimation and a learned interaction prior to recover 4D hand motion with a moving camera and interacting hands. Another paradigm, HaPTIC\cite{haptic} and OmniHands\cite{omnihands} are feed-forward methods. They extend HaMeR to process multiple frames jointly with cross-view self-attention and global cross-attention, and predict metric depth change to output coherent 4D hand trajectories. As HaPTIC showed promising performance in global trajectory accuracy, our system uses it to initialize the entire sequences, except the grasping frame, before optimizing for metric depths and hand-object interaction.

\subsection{Hand-Object Pose Estimation}
Hand-object pose estimation goes beyond hand-only modeling by additionally recovering object pose and shape, and enforcing physically plausible hand-object interactions. Prior work spans several directions. Template-based pipelines\cite{get_a_grip,arctic,artiboost,HOMAN,behave,cpf,Semi-Hand-Object,h+o,phosa} often assume a known object template, then estimate 6D object and human pose, e.g., HOMAN\cite{HOMAN} fits pose with reprojection and contact terms given the ground truth 3D CAD models. However, template-based pipelines hinge on object model availability thus are hard to scale in the wild. On the other hand, template-free methods utilize multi-view cues or data-driven priors to reconstruct unseen objects from RBG videos\cite{magichoi,hhor,diffhoi,hold,G-HOP,MCC-HO} or an RBG image\cite{gsdf,alignsdf,moho,ihoi}. DiffHOI\cite{diffhoi} leverages both by optimizing an object SDF and hand mesh over a video with multi-view reprojection from original views and a diffusion prior. DiffHOI requires category level hand-object supervision, while HOLD\cite{hold} is a category-agnostic pipeline that jointly reconstructs hands and unknown objects from a single monocular video by training a compositional implicit SDF model that disentangles hand/object geometry and uses contact/mask cues for pose refinement. However, these work require per-sequence test-time fitting, which is time consuming and prevents a single model across diverse objects, scenes, and actions. MCC-HO\cite{MCC-HO} presents a complementary retrieval-augmented feed-forward paradigm that infers a hand-conditioned object from RGB image, then retrieves a plausible 3D object via a text-to-3D generative model and rigidly aligns it over time, yielding temporally consistent trajectories for the hand-held object. Despite progress, these template-free pipelines often leave the non-holding sections unaddressed, which makes it hard to handle long-horizon manipulation. Across the directions, the surrounding scene is not constructed, and they often operate in hand-coordinates, which limits global depth/scale reasoning and downstream deployment. Our proposed system reconstructs a metrically aligned scene, uses a strong mesh prior for realistic object geometry, and optimizes a scene-aligned hand-object motion over the entire sequence, including non-holding and holding stages, while bridging gaps from detection failures via a human motion prior.
\section{Method}
\label{sec:method}

\begin{figure*}[t]
	\centering
	\includegraphics[width=1.0\textwidth]{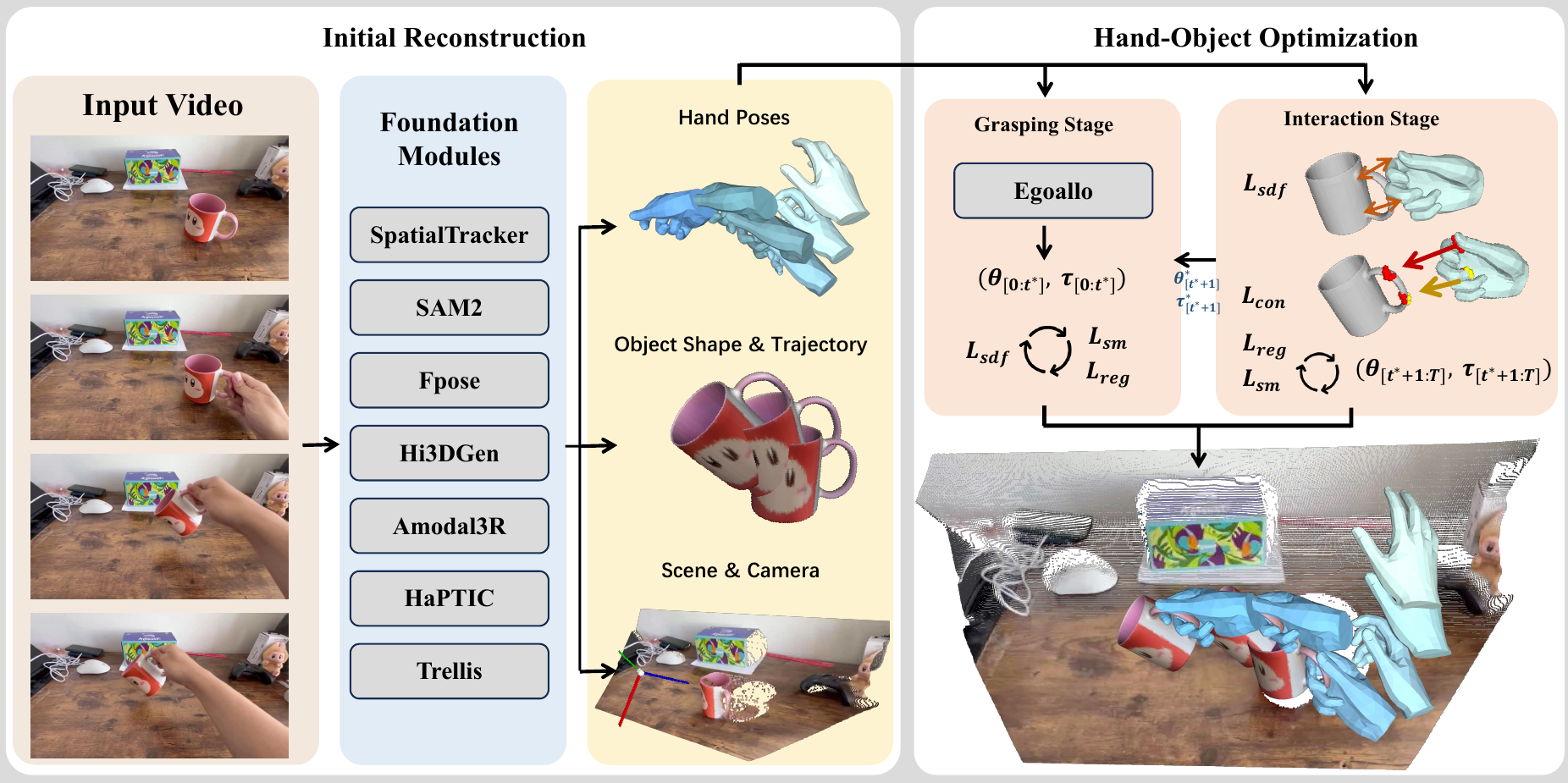}
        
	\vspace{-3mm}
    \caption{Overview of our framework. We first use foundation models to obtain the hand pose, object, and scene(Sec~\ref{sec:method_recon}). We then optimize hand–object interaction in the interaction stage by computing contact points and enforcing physical collision constraints(Sec~\ref{sec:inter_stage}). Finally, in the grasping optimization stage, we complete the motion using a human motion prior and further optimize the approaching and grasping phases (Sec~\ref{sec:grasp_stage}).}
	\label{fig:main}
\end{figure*}

\subsection{Problem formulation}
Given an RGB video $\mathbf{V} \in \mathbb{R}^{T\times H\times W\times 3}$ of human manipulating an object, our goal is to reconstruct the scene, textured object mesh and hand-object motion in the scene.  We represent scene geometry as a colored point cloud $\mathcal{P}$. The object $\Omega=(\mathcal{V}_o, \mathcal{F}_o)$ is modeled as a textured mesh with vertices $\mathcal{V}_o$ and faces $\mathcal{F}_o$. Object trajectories are parameterized by 6d poses $\mathcal{O} \in \mathbb{R}^{T\times 6}$. For hand representation, we adopt the MANO parametric model\cite{mano}, which includes subject-specific shape parameters $\beta \in \mathbb{R}^{10}$ and time-varying pose $\theta \in \mathbb{R}^{T\times 48}$ and global translation $\tau \in \mathbb{R}^{T\times 3}$. We denote the hand parameters as $\mathcal{H}=(\theta,\beta,\tau)$. We first describe how to reconstruct core elements as initialization using existing priors  in Section~\ref{sec:method_recon}, then explain how our optimization integrates these elements into a consistent 3D world in Section~\ref{sec:method_optim}.

\subsection{Initial Reconstructions}
\label{sec:method_recon}
For the scene and object components, we follow OnePoseviaGen\cite{oneposegen} to reconstruct the scene, the object in the scene, and the object's 6D pose trajectory. For hand pose, we adopt HaPTIC\cite{haptic} as the baseline to extract the hand trajectory. We now describe each component in detail. 

Before all the components, we first run SAM2\cite{sam2} on the entire video to get the consistent masks of the object. \\
\textbf{Scene Reconstruction.}
We reconstruct the scene by running SpatialTrackerV2\cite{spatialtrackerv2} to estimate video-consistent depth maps together with camera intrinsics and extrinsics. A pixel $(u,v)$ with depth $D_t(u,v)$ is back-projected to a 3D world point using the pinhole model (with world-to-camera extrinsics $[\mathbf{R}_t\,|\,\mathbf{t}_t]$ and intrinsics $\mathbf{K}$):
\[
P_t=\Bigl\{\,\mathbf{R}_t^{\!\top}\bigl(D_t(u,v)\,\mathbf{K}^{-1}[u,\,v,\,1]^\top-\mathbf{t}_t\bigr)\;\Big|\;(u,v)\in\Omega\,\Bigr\},
\]
where $\Omega$ denotes the set of valid pixels. Aggregating points over all $(u,v)\in\Omega$ for each frame $t$ forms a per-frame point cloud; concatenating across frames yields a dense, globally aligned scene reconstruction.
\\
\textbf{Textured Object Reconstruction.}
For textured object reconstruction, we send the first frame of the masked object image of the sequence to the State-of-the-art Image-to-3D generators. We route non-occluded objects to Hi3DGen\cite{hi3dgen} and occluded objects to Amodal3R\cite{amodal3r}.
% For the non-occluded object, we send the image to Hi3DGen\cite{hi3dgen}, and for the occluded object, we send to Amodal3R\cite{amodal3r}.
\\
\textbf{Object Pose Reconstruction.}
Given the image depth, camera intrinsics, and the generated object model, we further query Foundationpose\cite{foundationpose} to get the 6D object pose for each frames. We improve original Oneposeviagen by registering once on the first frame then tracking through the entire video, instead of querying 6D poses frame by frame.
\\
\textbf{Hand Pose Reconstruction.}
For the full-video hand reconstruction, we adopt HaPTIC\cite{haptic} as the backbone to produce sequences with coherent dynamics and smooth pose $\hat{\theta}$ and transitions $\hat{\tau}$. We perform a subsequent optimization to recover metric-accurate depths consistent with the RGB evidence, yielding scene-aligned hand trajectories.
\\

\begin{figure}

\begin{center}
\includegraphics[scale=0.42]{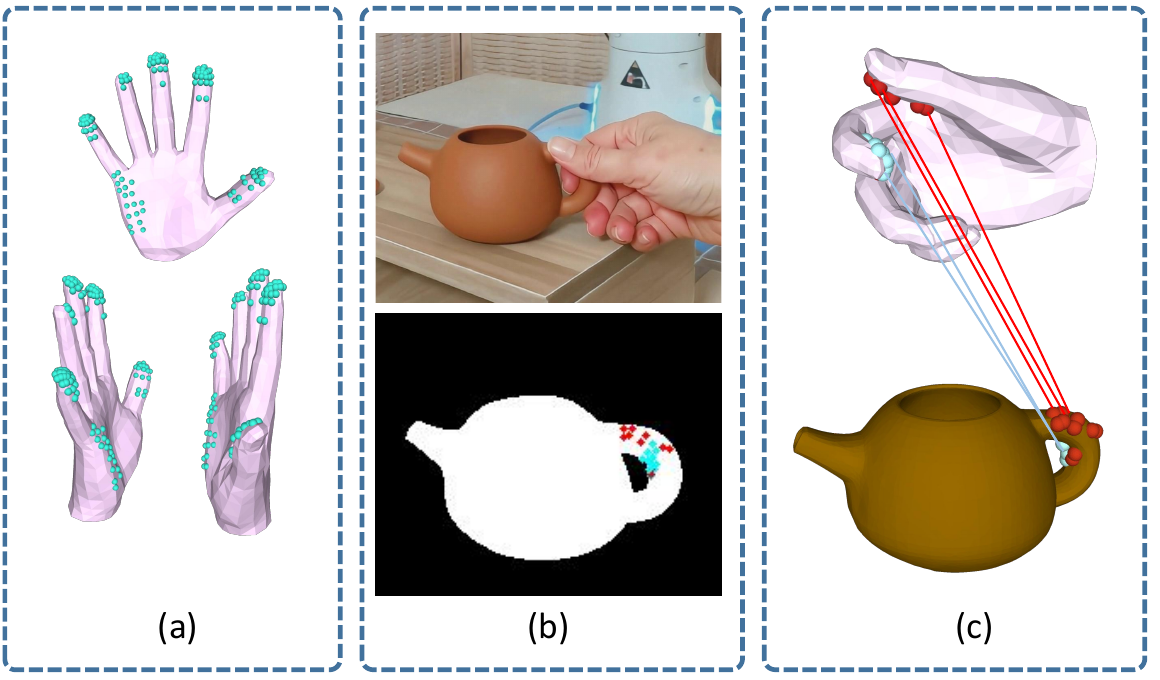}
\end{center}
\caption{(a) Contact candidates. (b) Top: input image. Bottom: the 2D projections of the contact candidates and their intersections with the object mask, where sky blue denotes back-surface contacts and red denotes front-surface contacts. (c) The correspondence between hand and object contact points.}
\label{fig:contact_candidates}
\end{figure}

\subsection{Hand-Object Motion Optimization}
\label{sec:method_optim}
\subsubsection{Two Stage Splitting}
We decompose the hand–object motion into two stages: grasping and interaction. The grasping stage spans from the hand’s approach to the onset of grasp. The interaction stage begins at grasp and covers the subsequent manipulation. This decomposition offers three benefits. 

(i) Objective disentanglement. In grasping, we target smooth, penetration-free translation. In interaction, we prioritize stable contact and object-conditioned motion. Jointly optimizing both introduces conflicting goals. (ii) Results from interaction optimization can be fedback to resolve the depth ambiguity that often remains in the grasping stage. (iii) Sequential extensibility. The two-stage design naturally handles multi-object sequences: approach (Stage 1), interact (Stage 2), then repeat Stage 1 for the next object.

We infer the grasping frame by the estimated object pose $\mathcal{O}=\{(o_0,r_0),\cdots,(o_T,r_T)\}$, where $o_i$ represents the object center translation at frame $i$ and $r_i$ represents the orientation. We set a translational threshold $\epsilon_o$ and a rotational threshold $\epsilon_r$. If, starting at time 
$t^\star$, the changes exceed these thresholds for $m$ consecutive frames, we regard 
$t^\star$ as the grasping frame. Let $\Delta o_k$ and $\Delta r_k$ denote the changes in position and rotation, the computation is given below:
\begin{align}
    t^\star_o =\inf\Bigl\{t \ge m :\min_{k\in\{t-m+1,\dots,t\}}
\bigl\| \Delta o_k\bigr\|_2>\epsilon_o \Bigr\}
\end{align}

\begin{align}
    t^\star_r=\inf\Bigl\{t \ge m :\min_{k\in\{t-m+1,\dots,t\}}\|\Delta r_k\|_2>\epsilon_r
\Bigr\}
% \\
% \qquad
% \theta_k
% =
% \arccos \Bigl(\frac{\mathbf{tr}(r_k r_{k-1}^\top)-1}{2}\Bigr).
\end{align}

Then we obtain $t^\star = \min (t^\star_o, t^\star_r)$.
We estimate the stationary pose of the object by averaging its pose in frames preceding grasping.
\begin{align}
    \hat{\mathcal{O}}_{[0:t^\star]} = \frac{1}{t^\star+1}\sum^{t^\star}_{i=0}\mathcal{O}_i
\end{align}

For the object poses after grasping, we leverage One-Euro Filter\cite{oneeurofilter} to obtain a smoother trajectory $\hat{\mathcal{O}}_{[t^\star+1:T]}$.The detailed procedures are in the supplementary material.
\\
\subsubsection{Interaction Stage Optimization} 
\label{sec:inter_stage}
During optimization, we first optimize the interaction stage to obtain a depth-consistent hand–object motion, then use the resulting depth estimates to optimize the grasping stage. In this stage, our goal is to find optimal hand poses $\theta^*$ and translations $\tau^*$ by several constraints:

\begin{align}
    (\theta^*,\tau^*)=\inf_{(\theta,\tau)}\{\lambda_1 L_{con} + \lambda_2 L_{sdf} + \lambda_3 L_{sm} +\lambda_4 L_{reg}\}
\end{align}
Next, we detail the semantics of these constraints.
\\
% \textbf{Constraints on 2D projection:} We first constrain the projection of the reconstructed hand mesh in the camera coordinate system. Haptic provides reasonably accurate 2D hand vertices locations $\mathcal{U}_h=\Pi_{K_h}\mathcal{V}_h$, but their depth in the scene is inaccurate because the camera intrinsics $K_h$ used by Haptic differ from those used in our scene reconstruction $K_s$ . Therefore, we optimize the hand’s 2D alignment by minimizing the distance between the 2D projections of the hand vertices under 
%  and the 2D points provided by Haptic.

% \begin{align}
%     L_{proj} = \|\Pi_{Ks}\mathcal{V}_h-\mathcal{U}_h\|_2
% \end{align}
 
% Once accurate 2D correspondences are available, one could estimate the 3D hand pose via methods such as SolvePnP; however, since this yields only an approximate solution, we use it solely as a constraint to refine the hand’s 2D position.
% \\

\begin{figure}
\includegraphics[scale=0.34]{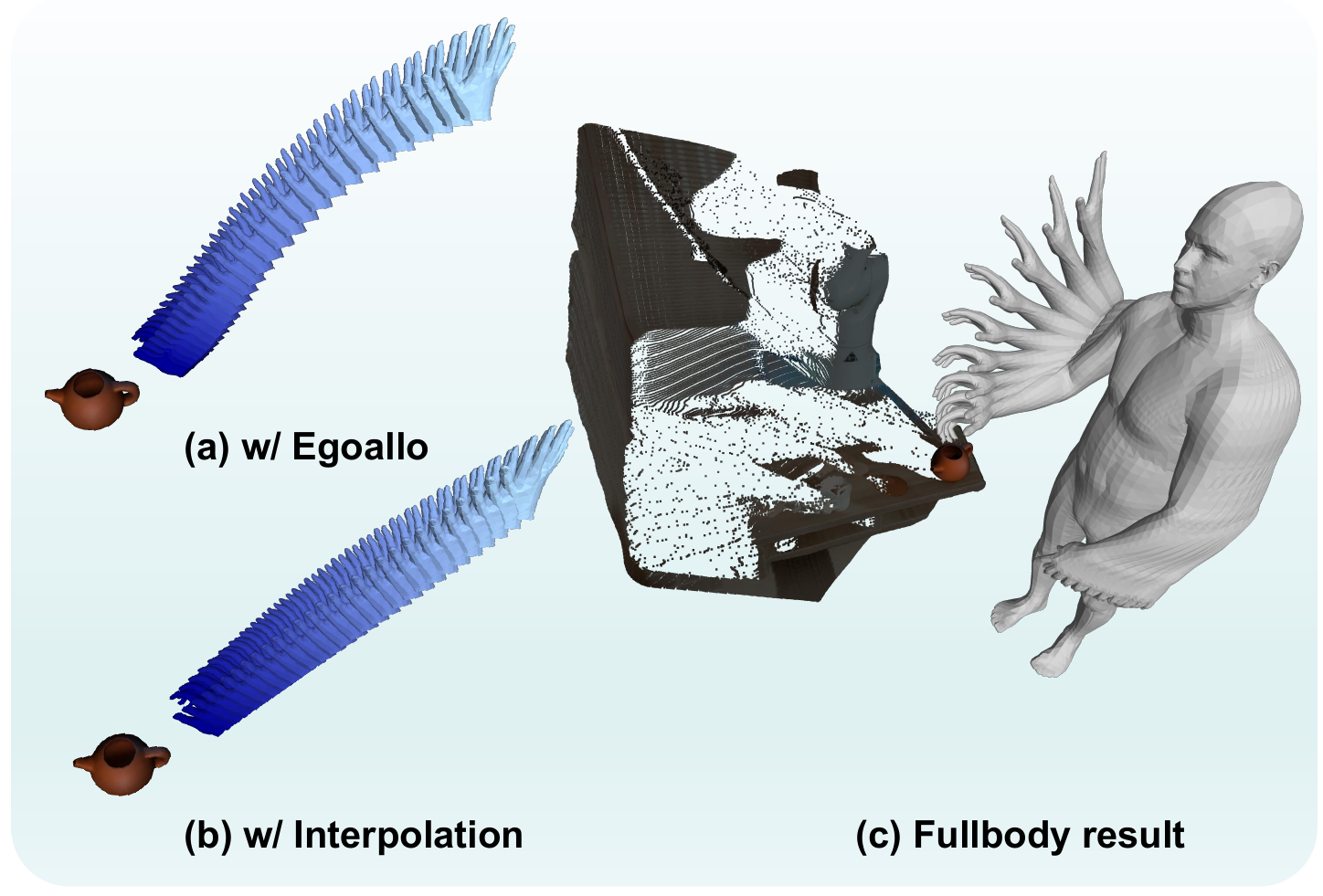}
\caption{Comparison of Egoallo generated motion, interpolation and fullbody motion. The initial hand pose is set to a half-raised posture. Darker color indicates progression over time.(a) shows the hand motion trajectory obtained with EgoAllo. (b) shows the trajectory produced by interpolation. (c) shows the full-body motion trajectory recovered by EgoAllo.}
\label{fig:egoallo}
\vspace{-3mm}
\end{figure}
\textbf{Constraints on contact Points.} To align the hand and object in 3D, we match corresponding contact points. In prior work\cite{liu2025easyhoi,magichoi}, sampling contact candidates from hand vertices is a choice shown to be efficient and effective. As demonstrated in Figure~\ref{fig:contact_candidates}, we sample 115 contact candidates $\mathcal{V}_c$ distributed in fingertips and treat these as potential contacts, shown in (a). Based on the intuition that vertices in contact in 3D also coincide in 2D, we project the candidate points onto the image plane using the camera intrinsics $K_s$ to obtain 2D contact candidates $\mathcal{U}_c=\Pi_{K_s}(\mathcal{V}_c$). 

With object mesh $\Omega$ and object pose $\hat{\mathcal{O}}$, we render a mask $M$ under the same camera intrinsics. Consequently, we compute the intersections of 2D contact candidates and object mask, which serve as the 2d contact points:
\begin{align}
    \mathcal{U}_\phi = M \cap \mathcal{U}_c
\end{align}

We proceed by pairing the 2D points with their 3D matches. Each 2D contact point corresponds to a hand vertex and a surface point on the object mesh. For hand-side contacts, we directly map the 2D contact set $\mathcal{U}_\phi$ to the corresponding 3D hand vertices $\mathcal{V}_\phi$ via their vertex indices. To find an object-side contact, we cast a ray from the camera through the 2D pixel, the contact lies at the intersection of this ray with the object surface. For a watertight object, the ray intersects the surface an even number of times, so we must decide whether the contact occurs on the front or back surface. 

Using MANO\cite{mano}, we obtain per-vertex normals $\mathcal{N}_h$ on the posed hand. Thus, for each 2D index, the corresponding hand-vertex normal is $\mathcal{N}_\phi$. With the viewing direction defined as the positive z-axis, vertices whose normal satisfies $z>0$ face away from the camera and are classified as back-surface hand contacts $\mathcal{V}_b$, whereas vertices whose normal satisfies $z < 0$ face toward the camera and are classified as front-surface hand contacts $\mathcal{V}_f$. For $\mathcal{V}_f$, their corresponding contact points on the object $O_f$ lie in the first intersections of the rays on the object surface. For $\mathcal{V}_b$, their contact points on the object $O_b$ are in the last intersections. %We denote the corresponding front points on object as $O_b$, and back points as $O_b$.% 
Then we define the contact loss as a distance loss:
\begin{align}
L_{\text{con}}
= \frac{1}{|\mathcal{V}_b|}\sum_{i=1}^{|\mathcal{V}_b|}\left\|\mathbf{v}_b^{(i)}-\mathbf{o}_b^{(i)}\right\|_2
\;+\;
\frac{1}{|\mathcal{V}_f|}\sum_{i=1}^{|\mathcal{V}_f|}\left\|\mathbf{v}_f^{(i)}-\mathbf{o}_f^{(i)}\right\|_2
\end{align}

\textbf{Constraints on Penetration.} 
Achieving a physically plausible grasp requires preventing finger–object interpenetration. We enforce a hand–object collision constraint using an SDF loss. Specifically, given a watertight object, we compute its signed distance field and denote the signed distance of hand vertex $v_i$ to the object surface by $d(v_i)$. The distance is positive when the vertex is outside the surface, and negative if inside. The penetration loss can be computed by:
\begin{align}
    L_{sdf}=\sum_{v\in \mathcal{V}_h} d(v)
\end{align}
where $\mathcal{V}_h$ refers to hand vertices.
\\

\textbf{Constraints on Motion Smoothness.}
Since hand pose and position should evolve smoothly throughout the motion, we enforce a smoothness loss. Specifically, we regularize temporal changes in the hand pose $\theta$ and the trajectories of the fingertip joints $\mathcal{J}^{tip}$:
\begin{align}
    L_{sm}=\frac{1}{T-t^\star-2}\sum_{t=t^\star+1}^{T-1}(\|\Delta\theta_{[t]}\| + \frac{1}{|\mathcal{J}^{tip}|}\sum_{j\in \mathcal{J}^{tip}}\|\Delta j_{[t]}\|)
\end{align}
where $\Delta \theta_{[t]}$ is the rotational change from frame $t$ to $t+1$ and $\Delta j_{[t]}$ refers to the change in joint position.

\textbf{Constraints on Regularization.}
We aim to optimize physical contact while preserving the original hand pose as much as possible, so we add a regularization term that penalizes deviations from the foundation model’s pose, which imposes a larger penalty as the optimized pose drifts farther from this reference:
\begin{align}
    L_{reg}=\frac{1}{T-t^\star-1}\|\hat{\theta}_{[t^\star+1:T]} - \theta_{[t^\star+1:T]}\|
\end{align}
where $\hat{\theta}$ denotes the pose estimated by the hand-detection baseline.
\\

\begin{figure*}[ht]
	\centering
	\includegraphics[width=0.9\textwidth]{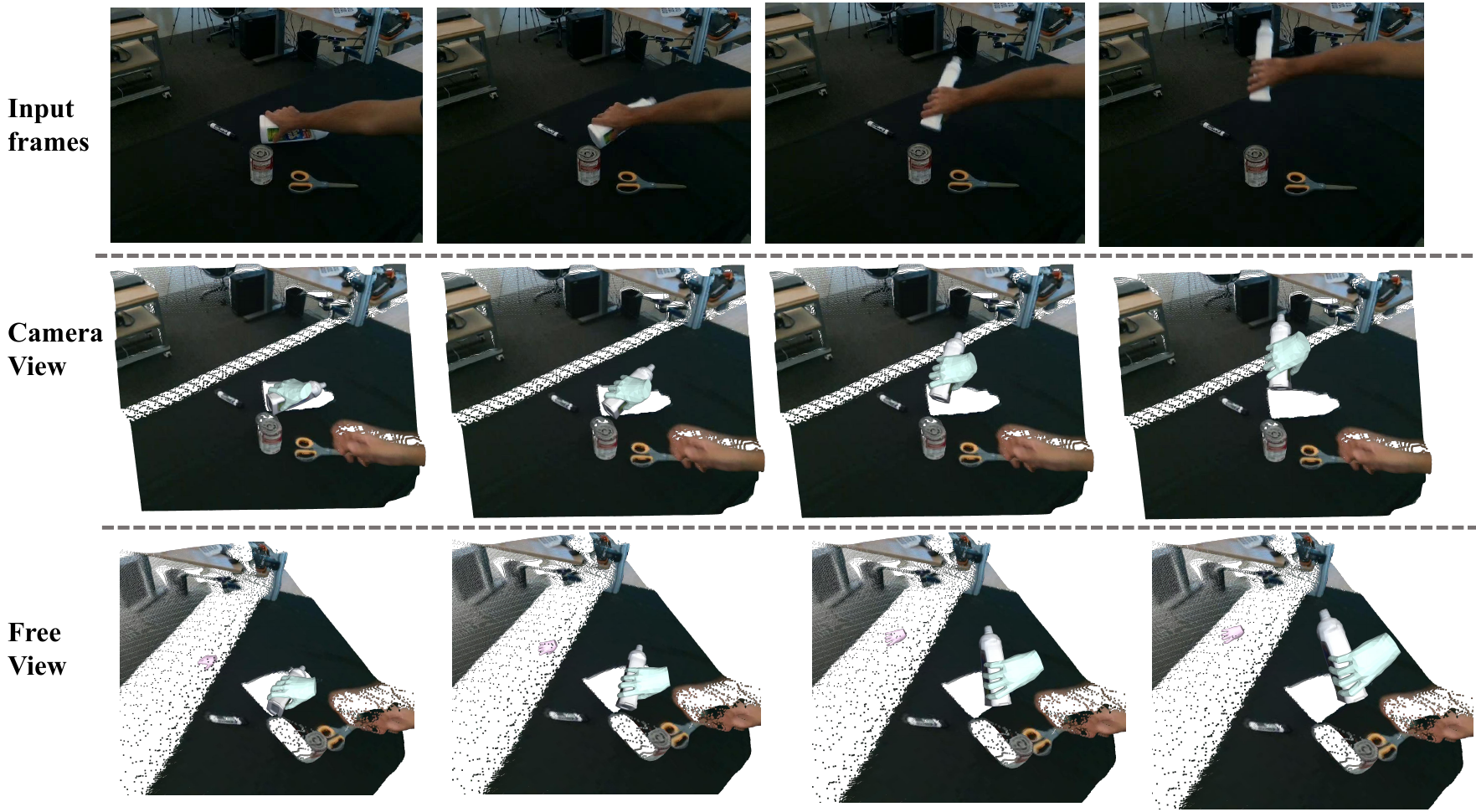}
    \vspace{-3mm}
    \caption{Qualitative results on DexYCB. We visualizes our reconstruction of the scene and object in 2 views, with the hand mesh shown in blue. In free view, the baseline hand reconstruction is overlaid in pink for comparison.
    }
	\label{fig:quali_dexycb}
    \vspace{-8pt}
\end{figure*}

\subsubsection{Grasping Stage Optimization}
\label{sec:grasp_stage}
\textbf{Grasp frame alignment.}
In this stage, we take the hand motion sequence $(\hat{\theta}_{[0:t^\star]},\hat{\tau}_{[0:t^\star]},\hat{\beta})$ extracted from Section~\ref{sec:method_recon} as the initialization.
After optimization of the interaction stage, we have already aligned the hand position in scene coordinate at the end of the grasping stage. Therefore, we initialize the target position $\hat{\tau}_{[t^\star]}$ for the grasping stage using the optimized $\tau^*_{[t^\star+1]}$ from the interaction stage. 
\begin{align}
    \hat{\tau}_{[0:t^\star]} \leftarrow \hat{\tau}_{[0:t^\star]} - \hat{\tau}_{[t^\star]}+\tau^*_{[t^{\star}+1]}
\end{align}

\textbf{Motion completion via human motion prior.}
\label{sec:method_egoallo}
During the approaching stage, the hand often moves from outside the camera’s field of view into the visible region. Since our hand detection algorithm is only effective in the hand visible frames $\mathcal{I}$, only initial pose and translation in these frames $\{\hat{\theta}_{[i]},\hat{\tau}_{[i]}\}_{i\in \mathcal{I}}$ are reliable, which requires us to complete the full hand motion. For the unseen segments outside the field of view, we aim to generate a complete approaching motion starting from a given initial pose. We require this motion to be smooth, kinematically plausible, and consistent with the valid poses at visible frames. To this end, we use Egoallo~\cite{yi2025egoallo}, a human motion prior to complete the motion using the guidance we have:
\begin{align}
    (\tilde{\theta},\tilde{\tau})=\pi(\{\hat{\theta}_{[i]},\hat{\tau}_{[i]}\}_{i\in \mathcal{I}})
\end{align}
By this prior, we achieve a complete approaching trajectory. In Figure~\ref{fig:egoallo}, we compare the motion trajectories generated by the guided motion prior with those obtained by interpolation and with real human motion. A thorough analysis of the effectiveness of this motion prior is provided in Section~\ref{sec:egoallo_effect}.

\textbf{Grasping motion optimization.}
We then use $(\tilde{\theta},\tilde{\tau})$ as the initialization for the subsequent optimization.
Our goal in this stage is to smoothen the approaching and grasping motion as well as avoiding finger–object interpenetration. We utilize smoothness loss to make the grasp pose coincide with the first frame of the second stage and further regularize the entire motion to be smoother. To prevent mesh interpenetration, we also impose an SDF loss as a constraint. The overall optimization loss for this phase can be represented:
\begin{align}
    L_{grasp}=\lambda_5 L_{sdf} + \lambda_6 L_{sm} +\lambda_7 L_{reg}
\end{align}

By optimizing with respect to the above constraints, we obtain $(\theta^*_{[0:t^\star]},\tau^*_{[0:t^\star]})$. We discuss the details of the parameter settings in the supplementary material.

\section{Experiments}
\label{sec:experiments}

\begin{figure*}[ht]
	\centering
	\includegraphics[width=1\textwidth]{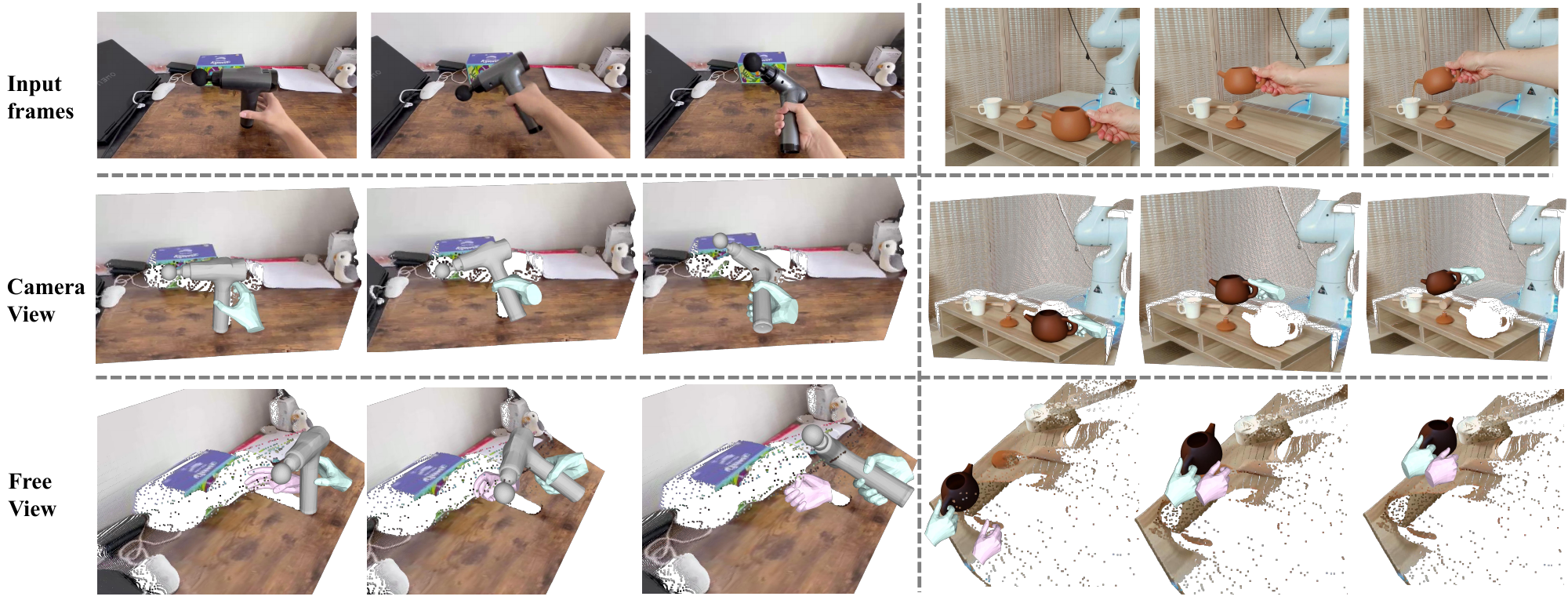}
    \vspace{-2mm}
    \caption{Qualitative results on in-the-wild videos. We evaluate on videos involving everyday object manipulation, including translating objects to target locations and rotating object orientations. We present results across multiple viewpoints and time instants, and compare them against the baseline.}
	\label{fig:quali_wild}
\end{figure*}

\begin{table*}[t]
  \centering
  \small
  \setlength{\tabcolsep}{4pt}
  \caption{Quantitative evaluation for in-scene object manipulation. We evaluate on two widely used HOI datasets HOI4D and DexYCB. We compare with the hand motion reconstruction baseline HaPTIC and the version it aligns the ground-truth wrist. We highlight the best values in bold.
}
  \label{tab:comparison}
  \begin{tabular}{l|l|ccccccc}
    \toprule
    \textbf{Dataset} & \textbf{Method}
      & MPJPE $\downarrow$ 
      & MRPE $\downarrow$ 
      & IV-mean $\downarrow$ 
      & IV-max $\downarrow$ 
      & ID-mean $\downarrow$ 
      & ID-max $\downarrow$ 
      & JM $\downarrow$ \\
    \midrule
    \textbf{HOI4D} 
      & HaPTIC\cite{haptic}            & 19.15 & 137.63 & -  & -   & -   & -    & \textbf{1.50} \\
      & HaPTIC$^\dagger$  & \textbf{19.15} & -   & 3.53 & 10.15 & 10.22 & 29.31 & 3.22 \\
      & Ours              & 20.48 & \textbf{64.10} & \textbf{0.46} & \textbf{4.34} & \textbf{3.68} & \textbf{24.24} & 2.38 \\
    \midrule
    \textbf{DexYCB} 
      & HaPTIC            & 7.88  & 150.72 & -  & -   & -   & -    & 0.82 \\
      & HaPTIC$^\dagger$  & 7.88  & -    & 8.91 & 14.25 & 15.14 & 35.69 & \textbf{0.71} \\
      & Ours              & \textbf{7.67} & \textbf{45.71} & \textbf{0.14} & \textbf{0.98} & \textbf{7.72} & \textbf{35.45} & 1.96 \\
    \bottomrule
  \end{tabular}
  \vspace{-8pt}
\end{table*}

To validate the effectiveness of our method, we evaluate it on two widely used hand-object datasets: DexYCB~\cite{chao2021dexycb} and HOI4D~\cite{liu2022hoi4d}. Since no existing methods address in-scene object manipulation, we compare our approach against our baseline and its global-aligned version.
\subsection{Datasets}
\textbf{DexYCB}\cite{chao2021dexycb} consists of videos in which 20 distinct YCB objects are grasped by hand, captured from 10 different viewpoints. Since our method is template-free, to reconstruct the object we filter out sequences where the target object is occluded by other objects retaining only those where it remains continuously visible and unobstructed. From this subset, we sampled 120 sequences, with an average of 52.11 frames per sequence. 
\\
\textbf{HOI4D}\cite{liu2022hoi4d} is a hand–object interaction video dataset captured with head-mounted cameras, containing both rigid and articulated objects. We evaluate on the rigid subset: “ToyCar,” “Mug,” “Bottle,” “Bowl,” “Kettle,” and “Knife.” We split each action segment within a video into a separate sequence. Similarly, we sampled 120 sequences  from both the training and test splits for evaluation,  with an average of 133.52 frames per sequence.

\subsection{Evaluation Metrics}
We evaluate both reconstruction accuracy and physical plausibility. Pose accuracy is measured by Mean Per Joint Position Error (MPJPE), and motion accuracy by Mean Root Position Error (MRPE), defined as the mean distance between predicted and ground-truth wrist positions over the sequence after aligning the first frame. Hand–object contact is assessed using interpenetration Volume (IV) and interpenetration Depth (ID), for which we report mean and maximum values, and motion smoothness is quantified by jerk magnitude (JM). MPJPE and MRPE are computed only on frames where the hand is visible, while IV, ID, and JM are computed over all frames in the interaction stage.

\subsection{Comparison Results}

\begin{table*}[!htbp]
 \centering
 \small
\caption{We ablate our components on DexYCB relative to the base model, reporting MPJPE, MRPE, and physics-based measures under different loss configurations. Row 1 is the unoptimized baseline, row 2 adds a contact loss with translation-only optimization, and rows 3–7 jointly optimize translation and pose with progressively richer loss combinations.}
\label{tab:dex_abl}
\begin{tabular}{cccc|ccccccc}
\toprule
$L_{con}$ & $L_{sdf}$ & $L_{reg}$ & $L_{sm}$ & MPJPE~$\downarrow$ & MRPE~$\downarrow$   & IV-mean~$\downarrow$ & IV-max~$\downarrow$ & ID-mean~$\downarrow$ & ID-max~$\downarrow$ & JM~$\downarrow$ \\ 
\midrule
$\times$                             & $\times$   & $\times$       & $\times$      & 7.88  & 150.72 & -       & -      & -       & -      & 0.82\\
$\checkmark$                             & $\times$   & $\times$       & $\times$      & 7.88  & 53.91  & 5.57    & 7.13   & 13.42   & 36.12  & 0.99 \\
$\times$                             & $\checkmark$   & $\checkmark$       & $\checkmark$      & 7.83  & 149.67 & \textbf{0.01}    & \textbf{0.07}  & \textbf{0.36}    & \textbf{2.50}    & \textbf{0.67} \\
$\checkmark$                             & $\times$   & $\checkmark$       & $\checkmark$      & 7.52 & 43.98  & 6.71    & 8.15   & 14.83   & 39.75  & 1.36 \\
$\checkmark$                             & $\checkmark$   & $\times$       & $\checkmark$      & 27.04 & 50.11  & 0.19    & 1.07   & 5.14    & 30.77  & 1.51 \\
$\checkmark$                             & $\checkmark$   & $\checkmark$       & $\times$      & \textbf{7.49}  & \textbf{40.92}  & 0.08    & 0.74   & 5.20    & 28.93  & 9.53 \\
$\checkmark$                             & $\checkmark$   & $\checkmark$       & $\checkmark$      & 7.67  & 45.71  & 0.14    & 0.98   & 7.72    & 35.45  & 1.96\\
\bottomrule
\end{tabular}
\end{table*}

\textbf{Quantitative Resutls.} We primarily compare hand pose accuracy, trajectory deviation, and a suite of physics-based metrics for hand–object interaction on DexYCB\cite{chao2021dexycb} and HOI4D\cite{liu2022hoi4d}. Since there are currently no other hand–object interaction methods operating in the world coordinate frame, we compare against baseline approach HaPTIC, which reconstructs hand motion in world coordiante frame. The detailed results are presented in Table~\ref{tab:comparison}. Since HaPTIC does not reconstruct objects, physical metrics cannot be computed directly. To enable a more informative comparison, we report HaPTIC \dag, obtained by aligning each frame’s hand wrist to the ground-truth wrist position, and then computing the physical metrics between reconstructed hand and the ground-truth object.

From the table, we observe that HaPTIC reconstructs reasonably accurate hand poses, but depth ambiguity leads to large trajectory errors. The HaPTIC\dag results further show that even after aligning the wrist to the ground truth, uncertainty in object shape yields substantial physical errors in the grasp configuration. In our method, reconstructing the object’s 3D shape enables more faithful recovery of the hand grasp. On HOI4D, because the ground truth itself contains hand–object penetrations, enforcing our physical optimization slightly degrades MPJPE.
\\
\textbf{Qualitative Results.} We present qualitative results in Figure~\ref{fig:quali_dexycb} and Figure~\ref{fig:quali_wild} on DexYCB and in-the-wild videos, respectively, visualizing the reconstructed scene, object, and hand motion. For each example, we show the input video frame, the reconstruction from the camera view, and a second free-view rendering. We also demonstrate comparisons with our baseline. The results indicate that our method produces correct grasps and accurate world-coordinate hand–object motion trajectories, whereas the baseline, due to depth ambiguity, fails to interact properly with the object.

\subsection{Ablation Studies.}
\label{sec:ablation}
\begin{table}[t]
    \centering
    \vspace{-3pt}
    \caption{We report the discrepancy between the motions generated by Egoallo and their corresponding guidance. }
    \label{tab:ego_abl}
    \setlength{\tabcolsep}{4pt}  
    \small
    \begin{tabular}{lcc}
        \toprule
        Dataset & MPJPE  & G-MRPE  \\
        \midrule
        DexYCB\cite{chao2021dexycb} & 0.041 & 0.170 \\
        HOI4D\cite{liu2022hoi4d}  & 0.087 & 0.134 \\
        \bottomrule
    \end{tabular}
    \vspace{-8pt}
\end{table}
\textbf{Effectiveness of optimization components.} 

We conduct an ablation study on the optimization components using DexYCB. Table~\ref{tab:dex_abl} reports results under different combinations of constraints.

Since our alignment hinges on contact points, omitting the contact loss yields little to no hand–object contact, which in turn produces deceptively low interpenetration metrics. In row 2 we Enable the contact loss while optimizing translation only. The results already delivers a notable boost in trajectory accuracy, underscoring the value of contact constraints. Removing the SDF loss eliminates collision penalties: pose and trajectory errors decrease, but both penetration volume and depth increase markedly. The fifth row illustrates the role of the regularization restriction, without it, optimization drifts toward physically plausible poses while deviating from the input observations. Finally, ablating the smoothness term can improve per-frame accuracy, but it introduces pronounced jitter over the full sequence. In the last row, using all constraints yields the best overall performance.
\\
\textbf{Effectiveness of human motion prior.} 

We employ the human motion prior EgoAllo~\cite{yi2025egoallo} to complete the approaching trajectory in Section~\ref{sec:method_egoallo}. EgoAllo is a diffusion-based human motion prior that can generate smooth motion, allowing control over hand movements via guidance signals. We provide discrete hand poses and wrist positions as guidance to generate the entire motion sequence. 
\label{sec:egoallo_effect}
Figure~\ref{fig:egoallo} compares trajectories produced with the motion prior versus interpolation. Our known inputs include the hand’s positions and poses on visible frames and a prescribed initial pose. As shown in (a), EgoAllo yields a trajectory that follows plausible human kinematics, whereas (b) the interpolated result exhibits mechanically linear motion. In (c), we visualize EgoAllo’s full-body output, which further demonstrates the realism and consistency of the completed motion. 

We quantitatively compare the input guidance to the corresponding generated outputs: MPJPE evaluates pose deviation, and global MRPE measures trajectory deviation. As shown in Table~\ref{fig:egoallo}, the MPJPE error is less than 0.1 mm, indicating that, in practice, the guidance and the generated results are effectively identical. 
\section{Discussion}
\textbf{Limitations and future work.} Our method relies on the accuracy of scene segmentation and the subsequent object reconstruction. In low-light conditions or under severe motion blur, object reconstruction may fail, which in turn prevents us from identifying correct contact points in the interaction stage. In addition, we currently perform object reconstruction using only the first frame, whereas in principle we could exploit object observations from the entire video, which is a promising direction for future improvement.
\\
\textbf{Conclusion.} We presented the first system for in-scene object manipulation reconstruction. We jointly estimate the whole scene and object-hand motion in scene frame coordinate, obtain the approaching motion and the interaction motion separately through a two-stage optimization scheme, and achieve strong performance on in-the-wild videos.
{
    \small
    \bibliographystyle{ieeenat_fullname}
    \bibliography{main}
}

\clearpage
\setcounter{page}{1}
\maketitlesupplementary
\appendix

\section{Parameter Setting}
\label{sec:param_setting}
We run our method on an NVIDIA L40 GPU. In Grasping stage optimization, we optimize $2000$ steps with the learning rate $10^{-2}$. In Interaction optimization, we optimize $8000$ steps with the learning rate $5\times 10^{-3}$. The loss weights we set in our method are as follow: $\lambda_1=2,\lambda_2=5,\lambda_3=1,\lambda_4=5,\lambda_5=2,\lambda_6=1,\lambda_7=5$. The units of our metrics are: MPJPE($mm$), MRPE($mm$), IV($cm^3$), ID($mm$), JM($mm/s^3$).

\section{Object pose processing}
For object poses in grasping stage, we estimate the stationary pose by averaging its pose in frames preceding grasping:
\begin{align}
    \hat{\mathcal{O}}_{[0:t^\star]} = \frac{1}{t^\star+1}\sum^{t^\star}_{i=0}\mathcal{O}_i
\end{align}
For the interaction stage, we apply a One-Euro filter\cite{oneeurofilter} to smooth the motion sequence. We set the minimum cutoff frequency to $f_{\min} = 1.5$, the derivative cutoff frequency to $f_d = 1.0$, the speed coefficient to $\beta = 0.01$, and the sampling interval to $\Delta t = 15$. We define the smoothing factor as:

\begin{align}
    \alpha(f,\Delta t)=\frac{\Delta t}{\Delta t+\tfrac{1}{2\pi f}} 
\end{align}
Initializaiton:
\begin{align}
    \hat o_{t^\star}=o_{t^\star},\quad \hat g_{t^\star}=\mathbf{0}
\end{align}
For each $t^\star < k < T$, we first estimate the derivative:
\begin{align}
     g_k=&\frac{o_k-\hat o_{k-1}}{\Delta t_k}
\end{align}
Then low-pass filter the derivative:
\begin{align}
    \hat o_k=&\alpha(f_d,\Delta t_k)\,g_k+\bigl(1-\alpha(f_d,\Delta t_k)\bigr)\hat g_{k-1}
\end{align}
Compute the adaptive main cutoff frequency:
\begin{align}
    f_c^{(k)}=&f_{\min}+\beta\,\lVert \hat g_k\rVert_2
\end{align}
Finally output the update:
\begin{align}
    \hat o_k=&\alpha\!\bigl(f_c^{(k)},\Delta t_k\bigr)\,o_k+\Bigl(1-\alpha\!\bigl(f_c^{(k)},\Delta t_k)\Bigr)\hat o_{k-1}.
\end{align}

After applying the above filtering, we obtain $\hat {\mathcal{O}}_{[t^\star+1:T]}$.

\begin{figure}

\begin{center}
\includegraphics[scale=0.26]{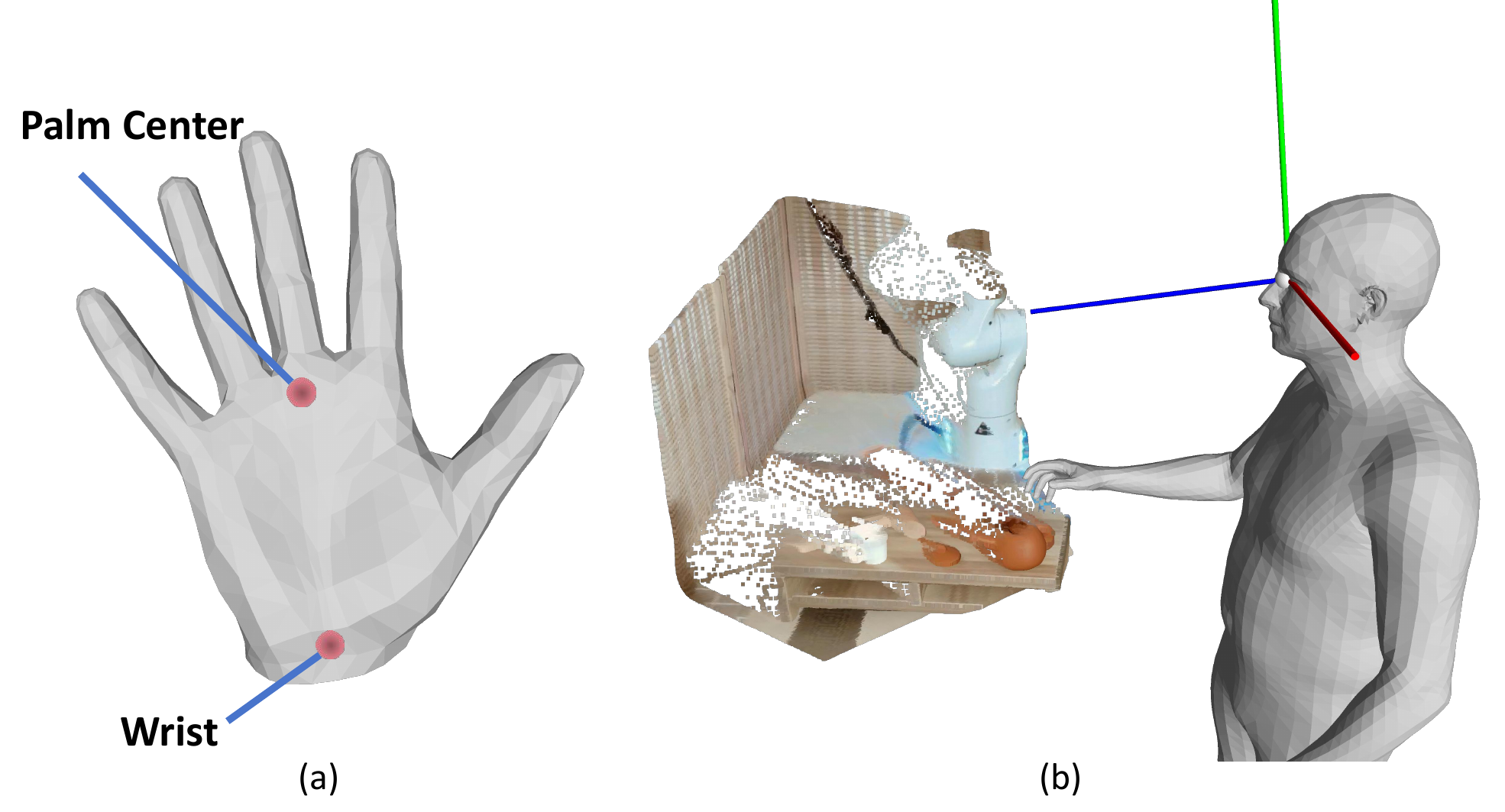}
\end{center}
\caption{We show (a) Positions of palm center and wrist on hand. (b)Input CPF coordinate and the corresponding generated SMPL-H human mesh.}
\label{fig:cpf_hand}
\end{figure}
\section{Application of Egoallo}
 Egoallo\cite{yi2025egoallo} takes a sequence of central pupil frame (CPF) as input and outputs a sequence of human motion represented by SMPL-H\cite{loper2023smpl,mano}, which aligns with the CPF.

We initialize the CPF using the camera pose $[\mathbf{R}_t\,|\,\mathbf{t}_t]$ predicted by SpatialTrackerV2\cite{spatialtrackerv2}, which roughly corresponds to a human viewing the object. However, since the manipulating hand is not always aligned with the camera’s egocentric viewpoint, we further adjust the CPF orientation and position according to the hand direction, ensuring that the resulting input corresponds to a natural human pose.

% Our world coordinate system is defined as an object-centered frame with the $z$-axis pointing upward. To obtain a natural body pose and position, we derive the body orientation from the grasping hand pose. Specifically, we take the "Palm Center" keypoint and "Wrist" keypoints from MANO\cite{mano} joints as demonstrated in Figure \ref{fig:cpf_hand}(a) as anchors and cast a ray from wrist to palm center. We then project this ray onto the $xy$-plane and get its rotation angle $r_{hand}$ around $z$ axis. Identically, we project the $z$ axis on CPF frame, which is the eye sight, onto the $xy$-plane in world frame coordinate and get rotation angle $r_{cpf}$. Then we can calculate 
% \begin{align}
%     \Delta r = \text{min}(r_{hand} - r_{cpf},\quad \frac{\pi}{6})
% \end{align}
% We require the hand-pointing direction to be reasonably consistent with the viewing direction. Therefore, when the angular difference exceeds $30^\circ$, we rotate the CPF around the $z$-axis in the world coordinate system, centered at the object, with the rotation angle given by $\Delta r$. In Figure \ref{fig:cpf_hand}(b), we show the CPF adjusted according to the grasping hand position, along with the resulting human body pose and position.

Our world coordinate system is defined as an object-centered frame with the $z$-axis pointing upward. To obtain a natural body pose and position, we align the CPF orientation with the grasping hand. Specifically, we use the MANO~\cite{mano} "Palm Center" and "Wrist" joints as anchors (Figure~\ref{fig:cpf_hand}(a)) and cast a ray from the wrist to the palm center. We project this ray onto the $xy$-plane and compute its azimuth angle $r_{\text{hand}}$ around the $z$-axis. Similarly, we project the CPF viewing direction which is its local $z$-axis onto the $xy$-plane in the world frame and obtain $r_{\text{cpf}}$. We then compute the angular difference
\begin{align}
    \Delta r = \text{min}(r_{hand} - r_{cpf},\quad \frac{\pi}{6})
\end{align}

We expect that the hand-pointing direction does not deviate too much from the viewing direction (at most $30^\circ$). When this angular difference exceeds $30^\circ$, we rotate the CPF around the $z$-axis in the world coordinate system, centered at the object, by $\Delta r$. Figure~\ref{fig:cpf_hand}(b) shows the CPF after this adjustment, together with the resulting human body pose and position.

\end{document}